\documentclass[10pt, a4paper]{article}

\usepackage{lrec-coling2024} 

\usepackage{fontspec}

\usepackage{natbib}
\usepackage{multibib}
\makeatletter
\def\@mb@citenamelist{cite,citep,citet,citealp,citealt,citepalias,citetalias}
\makeatother
\newcites{languageresource}{~}

\usepackage{graphicx}
\usepackage{tabularx}
\usepackage{soul}
\usepackage{titlesec}
\titleformat{\section}{\normalfont\large\bfseries\center}{\thesection.}{1em}{}
\titleformat{\subsection}{\normalfont\SmallTitleFont\bfseries\raggedright}{\thesubsection.}{1em}{}
\titleformat{\subsubsection}{\normalfont\normalsize\bfseries\raggedright}{\thesubsubsection.}{1em}{}
\renewcommand\thesection{\arabic{section}}
\renewcommand\thesubsection{\thesection.\arabic{subsection}}
\renewcommand\thesubsubsection{\thesubsection.\arabic{subsubsection}}

\usepackage{xcolor}
\usepackage{hyperref}
 \definecolor{darkblue}{rgb}{0, 0, 0.5}
  \hypersetup{colorlinks=true, citecolor=darkblue, linkcolor=darkblue, urlcolor=darkblue}

\usepackage{xstring}

\usepackage{colortbl}
\usepackage{booktabs}
\usepackage{amsmath}
\usepackage{hyperref}
\usepackage{graphicx}
\usepackage{subfig}
\usepackage{nicematrix,tikz}
\usepackage{stackengine}
\usepackage{authblk}
\usepackage{enumitem,kantlipsum}
\usepackage{arydshln}
\usepackage{diagbox}
\usepackage{multirow}
\usepackage{soul}

\newcommand*{\belowrulesepcolor}[1]{%
  \noalign{%
    \kern-\belowrulesep
    \begingroup
      \color{#1}%
      \hrule height\belowrulesep
    \endgroup
  }%
}
\newcommand*{\aboverulesepcolor}[1]{%
  \noalign{%
    \begingroup
      \color{#1}%
      \hrule height\aboverulesep
    \endgroup
    \kern-\aboverulesep
  }%
}

\NewDocumentCommand{\notsignif}{}{\makebox[0pt][l]{$^\dagger$}}

\title{A Controlled Reevaluation of Coreference Resolution Models}

\name{Ian Porada$^*$, Xiyuan Zou$^*$, Jackie Chi Kit Cheung} 

\address{McGill University, Mila - Quebec AI Institute \\
         \{ian.porada, xiyuan.zou\}@mail.mcgill.ca, jackie.cheung@mcgill.ca\\}

\abstract{
All state-of-the-art coreference resolution (CR) models involve finetuning a pretrained language model.
Whether the superior performance of one CR model over another is due to the choice of language model or other factors, such as the task-specific architecture, is difficult or impossible to determine due to lack of a standardized experimental setup.
To resolve this ambiguity, we systematically evaluate five CR models and control for certain design decisions including the pretrained language model used by each.
When controlling for language model size, encoder-based CR models outperform more recent decoder-based models in terms of both accuracy and inference speed.
Surprisingly, among encoder-based CR models, more recent models are not always more accurate, and the oldest CR model that we test generalizes the best to out-of-domain textual genres.
We conclude that controlling for the choice of language model reduces most, but not all, of the increase in F1 score reported in the past five years.
\\
\newline  }

\begin{document}

\maketitleabstract

\def\thefootnote{*}\footnotetext{Equal contribution.}

\section{Introduction}
\label{sec:introduction}

Coreference resolution (CR) is the task of clustering spans of text in a discourse that refer to the same entity or event \citep{hobbs1978resolving, hobbs1979coherence}.

All state-of-the-art CR models incorporate supervised finetuning of a pretrained language model. See \citet{poesio2023computational} for a comprehensive survey. We study the question:

\vspace{0.5em}
\noindent
\textit{To what extent are recent improvements in CR due to use of a more powerful language model as opposed to other design decisions?}
\vspace{0.5em}

\noindent
To answer this question, we evaluate existing CR models, controlling for the language model used in each. We find that at comparable language model sizes, encoder-based CR models outperform decoder-based models in terms of accuracy, inference speed, and memory usage.

This finding holds when we scale encoder-based models to sizes larger than existing work. For instance, when scaled to 1.5B parameters the encoder-based LingMess model \citep{otmazgin-etal-2023-lingmess} achieves 82.5 CoNLL F1 score on the CoNLL-2012 Shared Task test set. This is the same score as the decoder-based ASP model at 11B parameters \citep{liu-etal-2022-autoregressive}.

When we control for additional factors, such as the search space of supervised training epochs, we find that the oldest model tested, C2F \citep{lee-etal-2018-higher}, performs competitively with all other encoder-based models. Furthermore, C2F generalizes better to out-of-domain textual genres than all models of comparable size.

\textit{Based on our results, we conclude that controlling for the choice of language model reduces most, but not all, of the increase in F1 score reported in the past five years. Many improvements in CR model accuracy in may therefore be attributable to use of a stronger language model as opposed to other design decisions.} This finding suggests that future proposals intended to improve CR accuracy should carefully consider possible causes in order to better understand to what extent improvements are attributable to proposed architectural changes.

The five CR models that we test in our experiments are: 1) C2F; 2) S2E \citep{kirstain-etal-2021-coreference}; 3) WLC \cite{dobrovolskii-2021-word}; 4) LingMess; and, 5) Link-Append \citep{bohnet-etal-2023-coreference}. See \S\ref{sec:models} for model details. We focus our evaluation on English-language, document-level, nominal-entity coreference resolution as formulated in OntoNotes \citep{hovy-etal-2006-ontonotes}. We additionally test generalization to the GAP \citep{webster-etal-2018-mind} and OntoGUM \citep{zhu-etal-2021-ontogum} corpora. See \S\ref{sec:datasets} for dataset details.

The main contributions of this work are: \textbf{1)} We reimplement five state-of-the-art CR models, reproducing the original, published results. \textbf{2)} We show that more recent CR models are not always more accurate when we control for the choice of language model and the hyperparameter search space. \textbf{3)} We scale up encoder-based CR models to 1.5B parameters and find that they outperform decoder-based models of comparable size.

\section{Related Work}
\label{sec:related-work}

Existing work has presented controlled comparisons of published models, notably in the case of language model pretraining \citep{melis2018on,nityasya-etal-2023-scientific}. While similar in motivation to our work, we focus specifically on CR, with emphasis on the choice of language model.

Previous work has attempted to better understand state-of-the-art CR models based on targeted error analyses \citep{stoyanov-etal-2009-conundrums,lu-ng-2020-conundrums} and model generalization across datasets \citep{toshniwal-etal-2021-generalization,zabokrtsky-etal-2022-findings,porada2023investigating}; however, limited work has systematically evaluated existing model design. \citet{martschat-strube-2015-latent} performed a comprehensive evaluation of CR models, evaluating pre-neural architectures. More recent work has focused on analyzing particular steps in CR model architecture design \citep{toshniwal-etal-2020-cross,xu-choi-2020-revealing,lai2022end}.

Many existing CR models include ablations allowing for pairwise comparison between certain models \citep{dobrovolskii-2021-word,liu-etal-2022-autoregressive}; however, our study provides novel insights in that we train all models with the same competitive language model, where possible. Our results reveal the accuracy of C2F to be underestimated in existing comparisons.


\section{Methods}
\label{sec:methods}

In this section, we introduce details of each model, the factors that we control for, and the datasets used for evaluation.

\subsection{Models}
\label{sec:models}

We focus our evaluation on four encoder-based models and one decoder-based model, all of which reported state-of-the-art accuracy at their respective time of publication.
%
%

\subsubsection{Encoder-based Models}
\label{sec:encoder-based-models}

\paragraph{C2F} The C2F (Coarse-to-Fine) model \citep{lee-etal-2018-higher} is an extension of the earlier E2E \citep{lee-etal-2017-end}. E2E functions by encoding spans of text as contextualized vectors, and then pairwise scoring these vector representations using a task-specific head.
%
C2F extends E2E with an intermediate bilinear scoring function that filters down the number of spans considered. \citet{lee-etal-2018-higher} also proposed higher-order inference (HOI) and originally evaluated both E2E+C2F and E2E+C2F+HOI. We specifically consider only E2E+C2F, which we refer to as the C2F model, because HOI is known to marginally affect performance~\citep{xu-choi-2020-revealing}. While the original C2F proposal did not finetune the language model encoder, we follow the hyperparameters of \citet{joshi-etal-2019-bert}, which was the first work to do so.

\paragraph{S2E} The S2E (Start-to-End) model \citep{kirstain-etal-2021-coreference} is based on C2F with the main distinction that span representations are created using only the embeddings of the first and last token in the span. By contrast, span representations in C2F are a weighted sum of all token embeddings. S2E was motivated as requiring less memory while maintaining accuracy.

\paragraph{WLC} The WLC model (Word-Level Coreference; \citealp{dobrovolskii-2021-word}) is also based on C2F. In WLC, individual tokens are scored as candidate mentions, each representing the headword of some span.
WLC was motivated as being faster than C2F while also more accurate when using a RoBERTa language model \citep{liu2019roberta}.

\paragraph{LingMess} \citet{otmazgin-etal-2023-lingmess} proposed LingMess as a direct extension of S2E by increasing the number of task-specific, span-scoring heads.
LingMess significantly improved accuracy when using a Longformer language model \citep{Beltagy2020Longformer} as the encoder.


\subsubsection{Decoder-based Models}
\label{sec:decoder-based-models}

Two competitive decoder-based CR models have been proposed in the literature: ASP \citep{liu-etal-2022-autoregressive} and Link-Append. Of these we evaluate Link-Append, which reported a higher F1 score on OntoNotes, although we compare against ASP using published numbers where possible.


\paragraph{Link-Append} \citet{bohnet-etal-2023-coreference} proposed Link-Append, a method to finetune a language model for CR purely as a sequence generation task. For each sentence in the input document, a language model is trained to output a string of all coreferring mentions in the sentence. When used to finetune the 13B parameter mT5 language model \citep{xue-etal-2021-mt5}, Link-Append demonstrated state-of-the-art accuracy.

\begin{table}[t]
    \small
    \centering
    \begin{tabular}{llccc}
        \toprule
        Model & Language Model & Prior & Ours \\
        \midrule
        C2F & BERT-large & 76.9 & 77.1 \\
        S2E & Longformer-large & 80.3 & 80.4 \\
        WLC & RoBERTa-large & 81.0 & 81.0 \\
        LingMess & Longformer-large & 81.4 & 81.4 \\
        Link-Append & mT5-XXL & 83.3 & 83.3 \\
        \bottomrule
    \end{tabular}
    \caption{\label{tab:reimplementation-comparison}
        Test set performance on OntoNotes (CoNLL-2012) as reported in prior work (``prior'') and for our reimplementation of the model (``ours''). For each model, we use the best configuration of the respective prior work.
        \vspace{-10pt}
    }
    
\end{table}

\subsubsection{Implementation}

We reimplement all models using Hugging Face's Transformers library \citep{wolf-etal-2020-transformers} following each model's original hyperparameters.\footnote{Our code is available at \url{https://github.com/ianporada/coref-reeval} and datasets at \url{https://huggingface.co/coref-data}} This standardized setting allows for comparison of empirical inference speed and memory usage. Additionally, we can easily control for factors such as the language model used, whereas existing implementations often require manually-run preprocessing steps. We base our reimplementation on the original model code when available.

We verify our reimplementation of each model by comparing against the model accuracy for the best published configurations (Table~\ref{tab:reimplementation-comparison}).
As we do not have the resources to train the Link-Append model at the original 13 billion parameter size, we verify our processing steps by running inference on the released model weights. We also clarify preprocessing details with the original authors.




\subsection{Considered Factors}
\label{sec:controlling-for-confounds}

Our evaluation tests whether improved accuracy of a CR model is due to the proposed model design versus other factors. The main factor that we consider is the language model used in the initialization of each CR model. More recent models have also made changes to hyperparameters, such as the number of supervised training epochs, which may be responsible for improvements in performance. To control for these factors, we evaluate all models when tuned over the same hyperparameter space, a superset of existing search spaces.

\subsubsection{Language Model}

\paragraph{Encoder} For encoder-based CR models, we train all models using the same competitive language model encoder, DeBERTa \citep{he2021deberta}, at base and large sizes. We choose DeBERTa due to its competitive performance on the SuperGLUE benchmark \citep{NEURIPS2019_4496bf24}.

\paragraph{Encoder v.s. Decoder}

We do not directly compare encoder versus decoder-based CR models using the exact same language model because these classes of CR models in turn rely on language models of different architectures. Instead, we control for language model size. We train the best performing encoder-based model, LingMess, using DeBERTa encoders ranging from 138M to 1.5B parameters. We then compare against Link-Append and ASP decoder-based CR models trained with language models ranging from 300M to 13B parameters.

\begin{table}[t]
    \small
    \centering
    \begin{tabular}{lccccc}
        \toprule
        \belowrulesepcolor{gray!10}
        \rowcolor{gray!10} \multicolumn{6}{c}{\bf DeBERTa-base (138M)} \\
        \aboverulesepcolor{gray!10}
        \midrule
        Method & ON & OG & GAP & Mem. & Time \\
        \midrule
        C2F & 77.3 & 63.5 & 77.4 & 6.2 & 71.6 \\
        S2E & 78.6 & 63.8 & 78.9 & \textbf{1.3} & \textbf{20.9} \\
        WLC & 79.1 & 64.1 & 78.7 & 1.4 & 25.9 \\
        LingMess & \textbf{79.5} & \textbf{64.3} & \textbf{79.4} & 1.7 & 54.7 \\
        \midrule
        \midrule
        \belowrulesepcolor{gray!10}
        \rowcolor{gray!10} \multicolumn{6}{c}{\bf DeBERTa-large (405M)} \\
        \aboverulesepcolor{gray!10}
        \midrule
        Method & ON & OG & GAP & Mem. & Time \\
        \midrule
        C2F & 80.8 & \textbf{66.9} & 79.1 & 8.7 & 129.3 \\
        S2E & 80.5 & 65.9 & \textbf{79.9} & \textbf{2.7} & \textbf{45.3} \\
        WLC & 81.0 & 66.2 & 79.1 & 2.8 & 47.4 \\
        LingMess & \textbf{81.7} & 66.4 & 79.4 & 3.1 & 81.6 \\
        \bottomrule
    \end{tabular}
    \caption{\label{tab:encoder-comparison}
        Dev. set performance using a DeBERTa encoder of the given size. ON and OG are evaluated by CoNLL F1 score, GAP by F1 score. Mem. is max memory at batch size of one (GB), and time is inference speed in ms/doc at max batch size, both calculated w.r.t. inference on ON using a single 80GB A100 GPU.
    }
    \vspace{-10pt}
\end{table}

\subsubsection{Hyperparameter Search Space}

\paragraph{Finetuning Epochs}
More recent models have been trained for longer: while \citet{joshi-etal-2019-bert} trained C2F for 20 epochs, LingMess has been trained for 129 epochs, and the 100k steps used to train Link-Append amounts to around 160 epochs. To control for this change as a cause of improved accuracy, we search over training epochs in $\{25, 50, 125\}$ for each model.

\paragraph{Task-specific Head Size}
Post-C2F, encoder-based CR models use a larger hidden-layer size for task-specific heads (\texttt{ffnn\_size}). To control for the possible impact of this increase, we search over \texttt{ffnn\_size} $\in \{1024, 2048, 3072, 4096\}$ for every encoder-based model. This search space includes all sizes used by existing models.

\paragraph{Input Size} At training time, C2F reduces the length of documents to a small, fixed number of sentences in order to reduce memory usage, whereas other models do not have this constraint. We remove this constraint by training C2F on the maximum document length that fits into memory.

\begin{table*}[th]
    \small
    \centering
    \begin{tabular}{llc @{\hskip 3em} ccc @{\hskip 3em} cc}
        \toprule
        Model & LM & Size & ON & OG & GAP & Mem. & Time \\
        \midrule
        C2F & \multirow{4}{*}{DeBERTa$_\text{L}$} & 452M & 81.2\notsignif & \textbf{66.9} & 78.1 & 8.7 & 129.3 \\
        S2E & & 465M & 80.5 & 65.9\notsignif & \textbf{79.9} & \textbf{2.7} & \textbf{45.3} \\
        WLC & & 664M & 81.1\notsignif & 65.8\notsignif & 78.8 & 2.8 & 47.4 \\
        LingMess & & 561M & \textbf{81.7} & 66.4 & 79.4 & 3.1 & 81.6 \\
        \midrule
        \multirow{2}{*}{Link-Append} & mT5$_\text{B}$ & 582M & 66.2 & 45.8 & 63.3 & 5.9 & 5.2$\mathrm{e}$4 \\
        & mT5$_\text{L}$ & 1.23B & 66.5 & 45.5 & 64.9 & 12.3 & 1.4$\mathrm{e}$5 \\
        \bottomrule
    \end{tabular}
    \caption{\label{tab:cr-models-comparison}
        Dev. set accuracy when controlling for all factors. ON and OG are evaluated by CoNLL F1 score, GAP by F1 score. Mem. is the max memory usage for inference at batch size one (GB), and time is inference speed (ms/doc) at max batch size, both w.r.t. ON inference using a single 80 GB A100 GPU.
        F1 scores without $\dagger$ are statistically significantly different from all other scores based on a permutation test with $\alpha=0.1$ and 10k permutations following \citet{chinchor-1995-statistical}.
        }
\end{table*}

\subsection{Datasets}
\label{sec:datasets}

\paragraph{OntoNotes (ON)} As in existing work, we train and evaluate all models on the coreference annotations in the English-language portion of the OntoNotes 5.0 dataset \citeplanguageresource{ontonotes5}. We specifically use the CoNLL-2012 Shared Task v4 dataset splits \citep{pradhan-etal-2012-conll}.
For evaluation, we use the official CoNLL-2012 scorer. The train/validation/test splits are 1940/343/348 document parts, respectively.

\paragraph{OntoGUM (OG)} We additionally consider generalization to out-of-domain genres by evaluating models on the OntoGUM (OG) dataset \citep{zhu-etal-2021-ontogum}. OG is composed of the coreference annotations in the English-language GUM corpus \citep{Zeldes2017} transformed heurstically to follow OntoNotes annotation guidelines \citep{zhu-etal-2021-anatomy}. When evaluating on OG, we always evaluate on all 213 documents in the dataset.

\paragraph{GAP} Finally, we evaluate models on the 2000-example validation set of the GAP dataset \citep{webster-etal-2018-mind} as a targeted evaluation of pronominal anaphora. GAP consists of pronouns in English Wikipedia annotated for coreference with respect to two preceding noun phrases. We score models using the official scorer.

\section{Experiments}
\label{sec:experiments}

We now present results controlling for language model size and hyperparameter search space. We find that controlling for these factors considerably narrows the gap in accuracy between models.


\paragraph{More Recent Encoder-based Models Are Not Always More Accurate}

First, we test all CR models in their original configuration, with the minimal change that we use a DeBERTa encoder (Table~\ref{tab:encoder-comparison}). When using DeBERTa-large, the difference in CoNLL F1 on OntoNotes between all models reduces to less than 1 point absolute, and older models (C2F and S2E) are most accurate out-of-domain. Every encoder-based model also performs at or better than its best reported configuration in the literature.

We also control for the hyperparameter search space by comparing all models in their best hyperparameter configuration in Table~\ref{tab:cr-models-comparison}. In this case, the difference in performance between the oldest C2F and most recent LingMess encoder models reduces to 0.5 points absolute on OntoNotes. For all models in Table~\ref{tab:cr-models-comparison}, full precision, recall, and test set results are available in our project's GitHub repository, as are human-readable model predictions in CoNLL format.

\paragraph{Encoder-based Models Are More Accurate than Decoder-based Models at Comparable Sizes}

Table~\ref{tab:cr-models-comparison} shows the accuracy and inference time results for all encoder-based models versus Link-Append trained at approximately the same size. At this size, encoder-based models are substantially both more accurate and faster than Link-Append.

\begin{figure}[ht]
    \centering
    \includegraphics[width=0.45\textwidth]{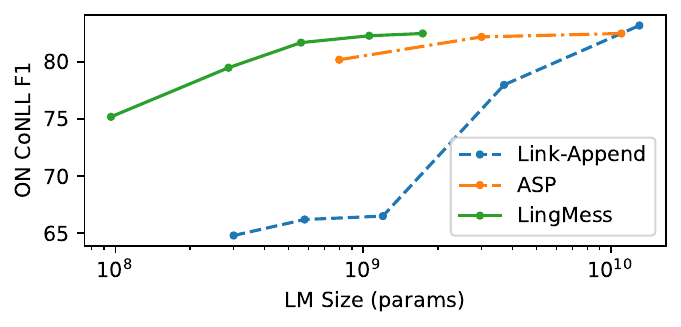}
    \caption{OntoNotes (ON) test set CoNLL F1 score for CR models trained with language models at multiple scales. Link-Append is trained with mT5, ASP with Flan-T5, and LingMess with DeBERTa. ASP scores, as well as the two largest Link-Append model scores, were reported by the respective authors. \citet{bernd-crac2023-keynote} also noted that Link-Append has an unexpected, apparently ``emergent'', scaling.}
    \label{fig:scaling}
\end{figure}

Accuracy of models trained over a range of sizes are shown in Figure~\ref{fig:scaling}. Up to the largest LingMess model trained, 1.5B parameters, LingMess outperforms decoder-based methods.

\begin{figure*}[ht]
    \centering
    \includegraphics[width=1.0\textwidth]{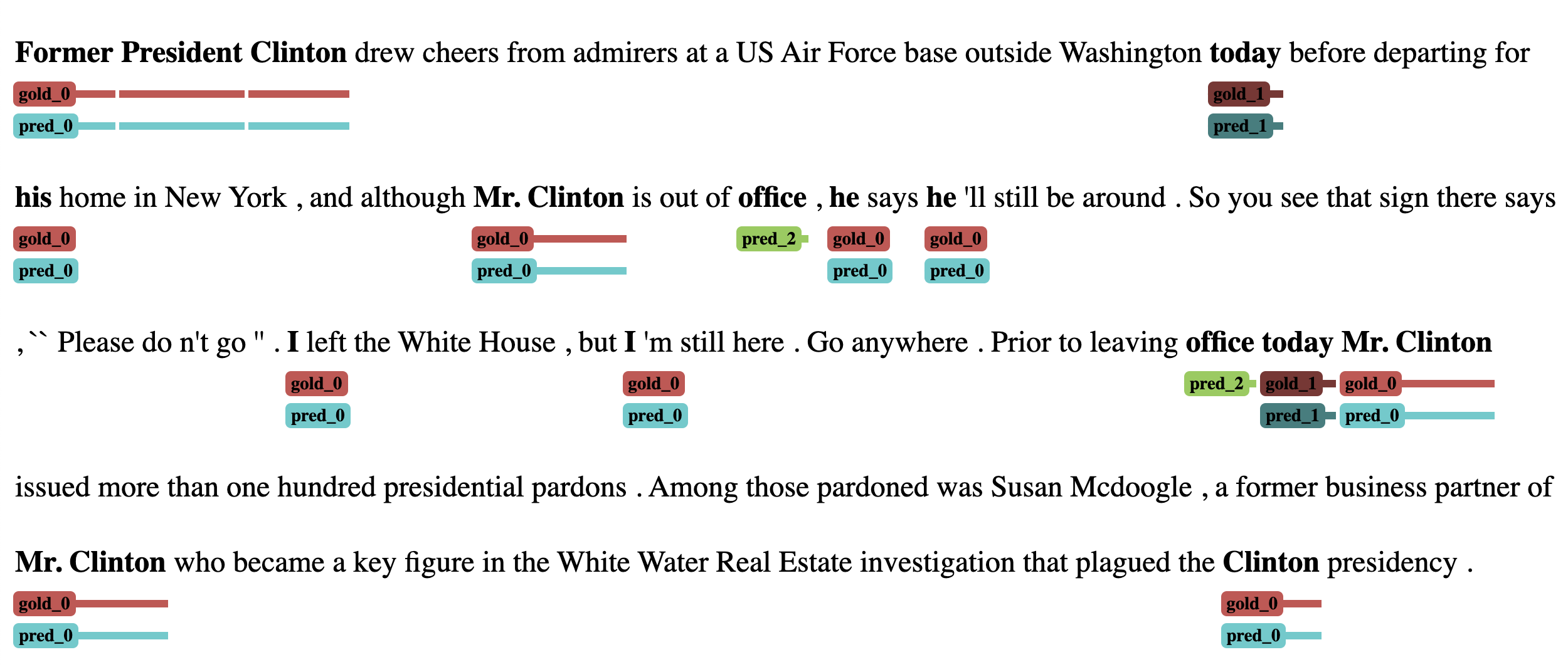}
    \caption{An example model output on the OntoNotes development set. ``Gold'' spans denote the ground-truth annotations, and ``pred'' spans denote predictions by the Link-Append mT5-XXL model.}
    \label{fig:example-output}
\end{figure*}

\subsection{Discussion}

Our results may be surprising because CR models more recent than C2F, such as WLC and S2E, are often reported to be more accurate than C2F when comparisons are presented \citep{chai-strube-2022-incorporating,liu-etal-2022-autoregressive,bohnet-etal-2023-coreference}; yet, we found the opposite to be true when controlling for the choice of language model. We suggest that future work explicitly considers these factors when presenting comparisons of CR architectures.

In some cases, less accurate models are substantially more memory or time efficient (Table~\ref{tab:cr-models-comparison}). While these dimensions have been considered in previous work \citep{kirstain-etal-2021-coreference}, even these efficient models have been motivated by their raw accuracy. To this end, we suggest more holistic evaluations of coreference beyond a single accuracy number, especially considering the fact that high OntoNotes F1 does not necessarily translate to good out-of-domain performance.

In our controlled comparison, the WLC model is never Pareto-optimal based on accuracy, runtime, and memory usage. All other encoder-based models are optimal in at least one of these three dimensions.

\section{Conclusion}
\label{sec:conclusion}

We reevaluate existing CR models, controlling for factors such as the choice of language model and number of training epochs. Among encoder-based models, the relative performance gap on OntoNotes narrows to 0.5 absolute CoNLL F1 score, and the oldest model tested (C2F) generalizes the best to out-of-domain textual genres. At all scales tested---up to 1.5B parameters---encoder-based CR models outperform decoder-based models of comparable model size; notably, LingMess trained with a 1.5B DeBERTa encoder has the same accuracy as decoder-based models over seven times larger in size.



\section*{Acknowledgements}

The authors acknowledge the material support of NVIDIA in the form of computational resources. Ian Porada is supported by a fellowship from the Fonds de recherche du Québec (FRQ). Xiyuan Zou is supported by a McGill Science Undergraduate Research Award (SURA) and FRQ - Bourse d’initiation à la recherche au 1er cycle (BIR). Jackie Chi Kit Cheung is supported by the Canada CIFAR AI Chair program. We thank Bernd Bohnet for help reimplementing the Link-Append model.

\section{Bibliographical References}\label{sec:reference}

\bibliographystyle{lrec-coling2024-natbib}
\bibliography{lrec-coling2024,anthology}

\section{Language Resource References}
\label{lr:ref}
\bibliographystylelanguageresource{lrec-coling2024-natbib}
\bibliographylanguageresource{languageresource}

\end{document}